\documentclass[11pt,letterpaper]{article}
\usepackage{emnlp2017}
\usepackage{times}
\usepackage{latexsym}
\usepackage{helvet}
\usepackage{courier,url}
\usepackage{graphicx,amsmath}
\usepackage{multirow,siunitx}
\usepackage[inline]{enumitem}
\usepackage{pbox}
\usepackage{subfig}
\usepackage{booktabs}
\newcolumntype{?}{!{\vrule width 1pt}}
\usepackage[inline]{enumitem} 

\usepackage{tikz}
\usetikzlibrary{positioning}

\usepackage{fixltx2e}
\usepackage{xcolor}

\emnlpfinalcopy

\def\emnlppaperid{730}

\def \eg {e.g., }

\def \etal {\ et al.\ }

\def \figname {Fig.}
\newcommand{\figref}[1]{\figname~\ref{#1}}
\def \Figname {Figure}
\newcommand{\Figref}[1]{\Figname~\ref{#1}}

\def \tabname {Table}
\newcommand{\tabref}[1]{\tabname~\ref{#1}}

\newcounter{typeno}

\usepackage{flushend}

\usepackage{color}

\usepackage[colorinlistoftodos]{todonotes}

\makeatletter
\def\Url@twoslashes{\mathchar`\/\@ifnextchar/{\kern-.2ejavascript:void(0);m}{}}
\g@addto@macro\UrlSpecials{\do\/{\Url@twoslashes}}
\makeatother

\title{Break it Down for Me:\\A Study in Automated Lyric Annotation}

    \author{
Lucas Sterckx\textsuperscript{*}, Jason Naradowsky\textsuperscript{$\dagger$}, Bill Byrne\textsuperscript{$\ddagger$}, \\ {\bf Thomas Demeester\textsuperscript{*}} \and  {\bf Chris Develder\textsuperscript{*}}  \\
    \textsuperscript{*} IDLab, Ghent University - imec \\
    {\tt{firstname.lastname@ugent.be}}\\
    \textsuperscript{$\dagger$} Language Technology Lab, DTAL, University of Cambridge \\ 
{\tt{ jrn39@cam.ac.uk}}\\
    \textsuperscript{$\ddagger$} Department of Engineering, University of Cambridge \\ 
{\tt{wjb31@cam.ac.uk}}
}
\date{}
\hypersetup{draft}
\begin{document}
\maketitle

\begin{abstract}
Comprehending lyrics, as found in songs and poems, can pose a challenge to human and machine readers alike.  This motivates the need for systems that can understand the ambiguity and jargon found in such creative texts, and provide commentary to aid readers in reaching the correct interpretation.

We introduce the task of automated lyric annotation (ALA).  Like text simplification, a goal of ALA is to rephrase the original text in a more easily understandable manner. However, in ALA the system must often include \emph{additional} information to clarify niche terminology and abstract concepts. To stimulate research on this task, we release a large collection of crowdsourced annotations for song lyrics. We analyze the performance of translation and retrieval models on this task, measuring performance with both automated and human evaluation. We find that each model captures a unique type of information important to the task.

\end{abstract}

\section{Introduction}
\label{sec:intro}
Song lyrics and poetry often make use of ambiguity, symbolism, irony, and other stylistic elements to evoke emotive responses.  These characteristics sometimes make it challenging to interpret obscure lyrics, especially for readers or listeners who are unfamiliar with the genre.  To address this problem,  several online lyric databases have been created where users can explain, contextualize, or discuss lyrics.  
Examples include MetroLyrics\footnote{{http://www.metrolyrics.com}} and Genius.com\footnote{{http://genius.com}}.  
We refer to such commentary as a lyric annotation (\Figref{fig:example}).

In this work we introduce the task of \emph{automated lyric annotation} (ALA).  
Compared to many traditional NLP systems, which are trained on newswire or similar text, an automated system capable of explaining abstract language, or finding alternative text expressions for slang (and other unknown terms) would exhibit a deeper understanding of the nuances of language.  As a result, research in this area may open the door to a variety of interesting use cases.  In addition to providing lyric annotations, such systems can lead to improved NLP analysis of informal text (blogs, social media, novels and other literary works of fiction), better handling of genres with heavy use of jargon (scientific texts, product manuals), and increased robustness to textual variety in more traditional NLP tasks and genres.

\begin{figure}[t]

\begin{tikzpicture}[auto,
    every edge/.append style = { <-, thick, >=stealth, gray,  line width = 1pt },
    every node/.append style = { align = center, minimum height = 10pt} ]
    \node[style= {font=\ttfamily\itshape}, text width = 7.6cm](c){How does it feel?\\
To be without a home\\
Like a complete unknown,};
  \node[below = .cm of c,style= {font=\ttfamily\bfseries\itshape}, text width = 6cm](d){ Like a rolling stone};
  \node [below = .4cm of d, style={fill= white,font=\bfseries\small}, text width = 7cm]  (D) {The proverb ``A rolling stone gathers no moss" refers to people who are always on the move, never putting down roots or accumulating responsibilities and cares.};
  \draw (D.north) edge (d.south);
\end{tikzpicture} 
   	\caption{A lyric annotation for ``Like A Rolling Stone" by Bob Dylan.}
	\label{fig:example}
\end{figure}

Our contributions are as follows:
\begin{enumerate}[noitemsep, topsep=0cm] 
\item To aid in the study of ALA we present a corpus of 803,720 crowdsourced lyric annotation pairs suitable for training models for this task.\footnote{To obtain the data collection please contact the first author of this paper.}
\item We present baseline systems using statistical machine translation (SMT), neural translation (Seq2Seq), and information retrieval. 
\item We establish an evaluation procedure which adopts measures from machine translation, paraphrase generation, and text simplification.  Evaluation is conducted using both human and automated means, which we perform and report across all baselines.
\end{enumerate}

\section{The Genius ALA Dataset}
\label{sec:data}
\label{sec:genius}
We collect a dataset of crowdsourced annotations, generated by users of the \textit{Genius} online lyric database.
For a given song, users can navigate to a particular stanza or line, view existing annotations for the target lyric, or provide their own annotation.  Discussion between users acts to improve annotation quality, as it does with other collaborative online databases like Wikipedia. 
This process is gamified: users earn \textit{IQ} points for producing high quality annotations. 

We collect 736,423 lyrics having a total 1,404,107 lyric annotation pairs from all subsections (rap, poetry, news, etc.) of Genius.  We limit the initial release of the annotation data to be English-only, and filter out non-English annotations using a pre-trained language identifier.  We also remove annotations which are solely links to external resources, and do not provide useful textual annotations.
This reduces the dataset to 803,720 lyric annotation pairs.
We list several properties of the collected dataset in \tabref{tab:data}. 

\begin{table}[t!]
\centering
  \begin{tabular}{l@{\hskip 0.8in}r}
  \toprule
\#  Lyric Annotation pairs &  803,720\\
$\oslash$ Tokens per Lyric & 15\\
$\oslash$ Tokens per Annotation& 43\\
  $|V_{\text{lyrics}}|$ & 124,022 \\
  $|V_{\text{annot}}|$ & 260,427 \\
  \bottomrule
\end{tabular}
\caption{Properties of gathered dataset ($V_{\text{lyrics}}$ and $V_{\text{annot}}$ denote the vocabulary for lyrics and annotations, $\oslash$ denotes the average amount). }
\label{tab:data}
\end{table}

\subsection{Context Independent Annotation}
\label{subsec:context}
\label{sec:baselines}
Mining annotations from a collaborative human-curated website presents additional challenges worth noting.
For instance, while we are able to generate large quantities of parallel text from Genius, users operate without a single, predefined and shared \emph{global} goal other than to maximize their own IQ points.  
As such, there is no motivation to provide annotations for a song in its entirety, or independent of previous annotations.

For this reason we distinguish between two types of annotations: \emph{context independent} (CI) annotations are independent of their surrounding context and can be interpreted without it, \eg explain specific metaphors or imagery or provide narrative while normalizing slang language. 
Contrastively, \emph{context sensitive} (CS) annotations provide broader context beyond the song lyric excerpt, \eg background information on the artist.

\par To estimate contribution from both types to the dataset, we sample 2,000 lyric annotation pairs and label them as either CI or CS.
Based on this sample, an estimated 34.8\% of all annotations is independent of context. 
\tabref{tab:data_examples} shows examples of both types.

\begin{table*}[h]
	\centering
  \footnotesize
  \begin{tabular}{@{}m{2.cm}ccm{10cm}@{}}
      \toprule
      \multirow{2}{*}{Type} & \% of &\multicolumn{2}{l}{\multirow{2}{*}{Examples}}\\
      & annotations \\
      \midrule
    CI & 34.8\%&[L]& \texttt{Gotta patch a lil kid tryna get at this cabbage}\\
  (Context& & [A]& He's trying to ignore the people trying to get at his money.\\ 
     independent)&& [L]&\texttt{You know it's beef when a smart brother gets stupid}\\
&& [A]& You know an argument is serious when an otherwise rational man loses rational. \\ 
  \midrule
 CS  & 65.2\% & [L]& \texttt{Cause we ain't break up, more like broke down}\\
(Context  && [A] &The song details Joe's break up with former girlfriend Esther.\\
sensitive)&& [L] &\texttt{If I quit this season, I still be the greatest, funk}\\
&& [A] &Kendrick has dropped two classic albums and pushed the artistic envelope further. \\
\bottomrule
\end{tabular}
\caption{Examples of context independent and dependent pairs of lyrics [L] and annotations [A].}
\label{tab:data_examples}
\end{table*}

While the goal of ALA is to generate annotations of all types, it is evident from our analysis that CS
annotations can not be generated by models trained solely on parallel text. That is, these annotations cannot be generated without background knowledge or added context. Therefore, in this preliminary work we focus on predicting CI lyric annotations.

\section{Baselines}
We experiment with three baseline models used for text simplification and paraphrase generation.
\begin{itemize}
\item \textbf{Statistical Machine Translation (SMT):}
One approach is to treat the task as one of translation, and to use established statistical machine translation (SMT) methods  ~\cite{DBLP:conf/emnlp/QuirkBD04} to produce them. We train a standard phrase-based SMT model to translate lyrics to annotations, using GIZA++~\cite{J03-1002} for word alignment and Moses~\cite{DBLP:conf/acl/KoehnHBCFBCSMZDBCH07} for phrasal alignment, training, and decoding. 

\item \textbf{Seq2Seq:} 
Sequence-to-sequence models~\cite{DBLP:conf/nips/SutskeverVL14}  offer an alternative to SMT systems, and have been applied successfully to a variety of tasks including machine translation. In Seq2Seq, a recurrent neural network (RNN) encodes the source sequence to a single vector representation.  A separate decoder RNN generates the translation conditioned on this representation of the source sequence's semantics.  We utilize Seq2Seq with attention~\cite{DBLP:journals/corr/BahdanauCB14}, which allows the model to additionally condition on tokens from the input sequence during decoding.

\item \textbf{Retrieval:}
In practice, similar lyrics may reappear in different contexts with exchangeable annotations. We treat the training corpus as a database of lyrics' excerpts with corresponding annotations, and at test time select the annotation assigned to the most similar lyric. 
This baseline is referred to as the \textit{retrieval} model. 
We use standard TF-IDF weighted cosine distance as similarity measure between lyrics' excerpts. 
\end{itemize}

\begin{table*}
\small
	\centering
  \begin{tabular}{@{}ll@{}}
 \toprule
 Lyric: & \texttt{and when I slept, I dream g's, son I need some}\\
Human: &  he dreams about large amounts of money because he is in need of it .\\
SMT: &      and when he slept, and he's dream, g's son, he needs some\\
Seq2Seq: &  when he sleeps , he dreams of getting money.\\
Retrieval: &   g's refer to the ghetto gangsters in london that the kid has become involved with\\
\midrule
Lyric: & \texttt{Mossberg pump I'm ridin' shotgun literally}\\
Human: &riding shotgun means to ride up front in the passenger seat\\
SMT: &   Mossberg pump - action shotgun, he's saying he's ridin' shotgun literally.\\
Seq2Seq: &      Mossberg is a firearms manufacturer. He is riding shotgun with his shotgun.\\
Retrieval: &   he fully respects the people mentioned earlier and is riding for them. \\
\midrule
Lyric: & \texttt{and my eyes red cause of all that haze}\\
Human: & his eyes turn red due to lack of hydration from the weed one smokes (in this case , smoking purple haze)\\
SMT: &     and his eyes are red because of all that haze.\\
Seq2Seq: &      his eyes are red because of all the weed he smokes.\\
Retrieval: &  his eyes are red as a sign of the fatigue of travel and , possibly , from drugs and alcohol .\\
\bottomrule
\end{tabular}
\caption{Lyrics excerpts with annotations from Genius (`Human') and automated annotators.
}
\label{tab:output_examples}
\end{table*}

\section{Evaluation}
\subsection{Data}
We evaluate automatic annotators on a selection of 354 CI annotations and partition the rest of the annotations into 2,000 instances for development and the full remainder for training. 
It is important to note that the annotations used for training and development include CI as well as CS annotations. 

Annotations often include multiple sentences or even paragraphs for a single lyrics excerpt (which does not include end marks), while machine translation models need aligned corpora at sentence level to perform well~\cite{DBLP:journals/tacl/XuNPCC16}.
We therefore transform training data by including each sentence from the annotation as a single training instance with the same lyric, resulting in a total of 1,813,350 sentence pairs. 

We use this collection of sentence pairs (denoted as \emph{sent.} in results) to train the SMT model. 
Seq2Seq models are trained using sentence pairs as well as full annotations.
Interestingly, techniques encouraging alignment by matching length and thresholding cosine distance between lyric and annotation did not improve performance during development. 

\begin{table*}[h]
	\centering
    \small
  \begin{tabular}{@{}lccrccccrcc@{}}
  \toprule
  & \multicolumn{2}{c}{Properties} && \multicolumn{4}{c}{Automated Evaluation}&& \multicolumn{2}{c}{Human Evaluation}\\
  \cmidrule{2-3} \cmidrule{5-8} \cmidrule{10-11}
&  {Length Ratio}&{Profanity/Tok.}&& {BLEU}& {iBLEU}&  {METEOR}&{SARI}   && {Fluency}& {Information} \\
\midrule
Human & 1.19 & 0.0027  && - & - & - & - & & 3.93& 3.53 \\
SMT (Sent.) & 1.23& 0.0068  && \underline{6.22} & 1.44 & \underline{12.20} & \underline{38.42} & & 3.82&  3.31  \\
Seq2Seq (Sent.)& 1.05 & 0.0023  && 5.33 & \underline{3.64} & 9.28 & 36.52 & & 3.76&  3.25 \\
Seq2Seq & 1.32& 0.0022  && 5.15 & 3.46 & 10.56 & 36.86 & & 3.83& \underline{3.34} \\
Retrieval & 1.18 & 0.0038 && 2.82  & 2.27 & 5.10 & 32.76 & & \underline{3.93} & 2.98  \\
\bottomrule
\end{tabular}
\caption{Quantitative evaluation of different automated annotators. }
\label{tab:scores_standard}
\end{table*}

\subsection{Measures}
For automated evaluation, we use measures commonly used to evaluate translation systems (BLEU, METEOR), paraphrase generation (iBLEU) and text simplification (SARI).

BLEU~\cite{DBLP:conf/acl/PapineniRWZ02} uses a modified form of precision to compare generated annotations against references from Genius.
METEOR~\cite{DBLP:journals/mt/LavieD09} is based on the harmonic mean of precision and recall and, along with exact word matching, includes stemming and synonymy matching.
iBLEU~\cite{DBLP:conf/acl/SunZ12} is an extension of the BLEU metric to measure diversity as well as adequacy of the annotation, iBLEU $= 0.9$ $\times$ BLEU(Annotation, Reference) $-$ 0.1~$\times$ BLEU(Annotation, Lyric).
SARI~\cite{DBLP:journals/tacl/XuNPCC16} measures precision and recall of words that are added, kept, or deleted separately and averages their arithmetic means.

We also measure quality by crowdsourcing ratings via the online platform CrowdFlower.\footnote{{https://www.crowdflower.com/}} We present collaborators  with a song lyric excerpt annotated with output from the annotation generators as well as a reference annotation from Genius. Collaborators assign a 5-point rating for \emph{Fluency} which rates the quality of the generated language, and \emph{Information} which measures the added clarification by the annotation, a key aspect of this task. For each lyric annotation pair, we gather ratings from three different collaborators and take the average. 

\subsection{Hyperparameters and Optimization}
Here we describe implementation and some of the optimizations used when training the models.

For Seq2Seq models, we use OpenNMT~\cite{2017opennmt} and optimize for perplexity on the development set. Vocabulary for both lyrics and annotations is reduced to the 50,000 most frequent tokens and are embedded in a 500-dimensional space.  

We use two layers of stacked bi-directional LSTMs with hidden states of 1024 dimensions. We regularize using dropout (keep probability of 0.7) and train using stochastic gradient descent with batches of 64 samples for 13 epochs.

The decoder of the SMT model is tuned for optimal BLEU scores on the development set using minimum error rate training~\cite{DBLP:journals/pbml/BertoldiHF09}.

\begin{figure}[h] 
\includegraphics[width=7.8cm]{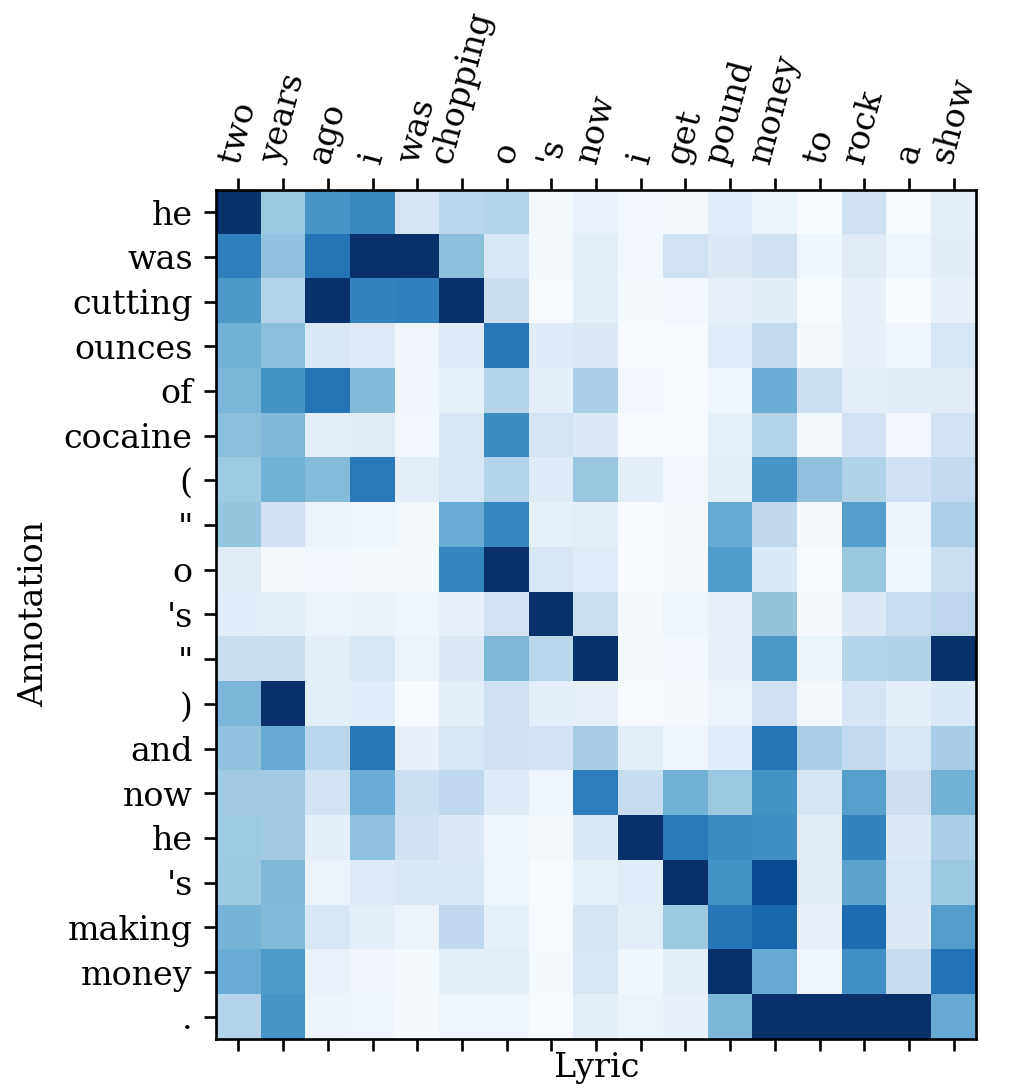}
\caption{Attention visualization of Seq2Seq models for ALA.}
\label{fig:att}
\end{figure}

\section{Results}

To measure agreement between collaborators, we compute the kappa statistic~\cite{fleiss1971measuring}. 
Kappa statistics for fluency and information are 0.05 and 0.07 respectively, which indicates low agreement. The task of evaluating lyric annotations was difficult for CrowdFlower collaborators as was apparent from their evaluation of the task.
For evaluation in future work, we recommend recruitment of expert collaborators familiar with the Genius platform and song lyrics. 

\tabref{tab:output_examples} shows examples of lyrics with annotations from Genius and those generated by baseline models. 

A notable observation is that translation models learn to take the role of narrator, as is common in CI annotations, and recognize slang language while simplifying it to more standard English.

Automatic and human evaluation scores are shown in \tabref{tab:scores_standard}. Next to evaluation metrics, we show two properties of automatically generated annotations; the average annotation length relative to the lyric and the occurrence of profanity per token in annotations, using a list of 343 swear words. 

The SMT model scores high on BLEU, METEOR and SARI but shows a large drop in performance for iBLEU, which penalizes lexical similarity between lyrics and generated annotations as apparent from the amount profanity remaining in the generated annotations.

Standard SMT rephrases the song lyric from a third person perspective but is conservative in lexical substitutions and keeps close to the grammar of the lyric. 
A more appropriate objective function for tuning the decoder which promotes lexical dissimilarity as done for paraphrase generation, would be beneficial for this approach.

Seq2Seq models generate annotations more dissimilar to the song lyric and obtain higher iBLEU and Information scores. 
To visualize some of the alignments learned by the translation models,~\figref{fig:att} shows word-by-word attention scores for a translation by the Seq2Seq model.

While the retrieval model obtains quality annotations when test lyrics are highly similar to lyrics from the training set, retrieved annotations are often unrelated to the test lyric or specific to the song lyric it is retrieved from. 

Out of the unsupervised metrics, METEOR obtained the highest Pearson correlation~\cite{pearson1895note} with human ratings for Information with a coefficient of 0.15.

\section{Related Work}
\label{sec:relatedwork}
Work on modeling of social annotations has mainly focused on the use of topic models~\cite{DBLP:conf/nips/IwataYU09,DBLP:conf/wsdm/DasBB14} in which annotations are assumed to originate from topics. They can be used as a preprocessing step in machine learning tasks such as text classification and image recognition but do not generate language as required in our ALA task.

\par Text simplification and paraphrase generation have been widely studied. Recent work has highlighted the need for large text collections~\cite{xu2015problems} as well as more appropriate evaluation measures~\cite{DBLP:journals/tacl/XuNPCC16,deltableu}. They indicated that especially informal language, with its high degree of lexical variation, \eg as used in social media or lyrics, poses serious challenges~\cite{Xu_gatheringand}.

\par Text generation for artistic purposes, such as poetry and lyrics, has been explored most commonly using templates and constraints~\cite{DBLP:conf/ecai/BarbieriPRE12}. In regard to rap lyrics, Wu \etal\shortcite{DBLP:conf/emnlp/WuASB13} present a system for rap lyric generation that produces a single line of lyrics that is meant to be a response to a single line of input. Most recent work is that of Zhang\etal\shortcite{DBLP:conf/emnlp/ZhangL14} and Potash\etal\shortcite{DBLP:conf/emnlp/PotashRR15}, who show the effectiveness of RNNs for the generation of poetry and lyrics. 

\par The task of annotating song lyrics is also related to metaphor processing. As annotators often explain metaphors used in song lyrics, the Genius dataset can serve as a resource to study computational modeling of metaphors~\cite{metaphorL10-1419}.

\section{Conclusion and Future Work}
\label{sec:conclusion}

We presented and released the Genius dataset to study the task of Automated Lyric Annotation. As a first investigation, we studied automatic generation of context independent annotations as machine translation and information retrieval. Our baseline system tests indicate that our corpus is suitable to train machine translation systems.

Standard SMT models are capable of rephrasing and simplifying song lyrics but tend to keep close to the structure of the song lyric. Seq2Seq models demonstrated potential to generate more fluent and informative text, dissimilar to the lyric.

A large fraction of the annotations is heavily based on context and background knowledge (CS), one of their most appealing aspects. 
As future work we suggest injection of structured and unstructured external knowledge~\cite{2016arXiv160800318A} and explicit modeling of references~\cite{2016arXiv161101628Y}.

\section*{Acknowledgments}
The authors wish to thank the anonymous reviewers for their helpful comments. This work was supported by the Research Foundation - Flanders (FWO) and the U.K. Engineering and Physical Sciences Research Council (EPSRC grant EP/L027623/1).

\bibliographystyle{emnlp_natbib}
\bibliography{emnlp2017}

\end{document}